\documentclass[11pt]{article}

\usepackage[final]{acl}
\usepackage{amsmath}
\usepackage{times}
\usepackage{latexsym}
\usepackage{multirow}
\usepackage[T1]{fontenc}

\usepackage[utf8]{inputenc}

\usepackage{microtype}

\usepackage{inconsolata}

\usepackage{graphicx}

\usepackage{booktabs}
\usepackage{threeparttable}
\title{Does Continued Pretraining on a Learner Corpus Improve Automated Essay Scoring on English Proficiency Tests? Evidence from EFCAMDAT}

\author{
\textbf{Duy Anh Nguyen\textsuperscript{1,2}} \\
\textsuperscript{1}University of Greenwich, London, United Kingdom \\
\textsuperscript{2}FPT University, Can Tho, Vietnam \\
\texttt{na6367q@gre.ac.uk} \\
\texttt{anhndgcc240003@gmail.com}
}

\begin{document}
\maketitle
\raggedbottom
\begin{abstract}
Automated Essay Scoring (AES) for English proficiency assessment increasingly relies on pretrained transformer models, yet these models are typically trained on general-domain English and may under-represent second-language learner writing. This study investigates whether domain-adaptive continued pretraining (DAPT) on a learner-writing corpus improves transformer-based AES for English proficiency assessment. We perform DAPT on BERT, RoBERTa, and DistilBERT using the EFCAMDAT corpus, then compare the adapted models with their original checkpoints on two English proficiency test datasets, FCE and IELTS, in both in-domain scoring and few-shot cross-dataset transfer. Full-corpus DAPT produces mixed effects across models, datasets, and metrics. Subsequent lexical and syntactic analyses suggest mismatches between EFCAMDAT and the downstream datasets in proficiency level, genre, and communicative purpose. We therefore repeat DAPT using proficiency-specific EFCAMDAT subsets across all three encoder architectures. Proficiency-specific DAPT frequently outperforms full-corpus DAPT and, in some settings, even the non-adapted baseline. Overall, continued pretraining on learner writing can improve in-domain AES, but its benefits depend on both the proficiency composition of the pretraining data and the underlying encoder architecture, and do not consistently extend to cross-test transfer.
\end{abstract}

\section{Introduction}
English proficiency assessment plays an important role in educational placement, certification, and language learning, with writing being a key component because it reflects a learner’s ability to organize ideas and communicate effectively in English. However, evaluating writing is resource-intensive and time-consuming, especially when large numbers of responses must be assessed consistently. Therefore, Automated Essay Scoring (AES) has clear practical value in English proficiency settings, where it can support efficient and timely evaluation of learner writing.

Despite this practical importance, AES for English proficiency writing remains challenging. Unlike many general essay datasets, English assessment essays are produced by L2 learners, which often contain distinctive lexical, grammatical, and discourse patterns associated with developing language ability. Recent transformer-based AES approaches commonly rely on general-domain pretrained language models \cite{qiu-etal-2024-large-language,schmalz-brutti-2021-automatic}. Such pretraining may provide useful knowledge of conventional English against which characteristics of learner writing can be distinguished. However, these models are not explicitly pretrained on the linguistic distribution of L2 writing, raising the question of whether additional pretraining on learner data can complement their existing representations and better adapt them to this setting.

Beyond the challenge of modeling learner writing itself, AES for English proficiency assessment must also contend with variation across tests. Currently, a wide range of standardized English proficiency tests are used worldwide, such as IELTS, TOEFL, and Cambridge English examinations, each of which differs in task type, prompt design, scoring criteria, and assessment context. This diversity creates a practical challenge for AES, as a model developed for one test may not perform reliably on another. Building a separate AES system for every individual test would be both time-consuming and inefficient, especially as new test formats and writing tasks continue to emerge. A more practical direction is therefore to develop an AES approach with stronger transferability, so that one model can generalize more effectively across different English proficiency assessments. 

This study investigates whether continued pretraining on an English learner corpus yields improved AES performance on English proficiency tests. We perform domain-adaptive pretraining on BERT, RoBERTa, and DistilBERT, using the EFCAMDAT corpus \cite{geertzen2014efcamdat, huang2017efcamdat_users} as pretraining data. Our study conducts two main experiments: baseline comparisons and cross-dataset transfer, using two different datasets for downstream scoring tasks.

We hypothesize that continued pretraining on a learner corpus can improve both in-domain and cross-dataset AES performance by adapting general-domain representations to the linguistic distribution of L2 learner writing represented in the downstream assessment data. Taken together, our research aims to answer the following questions:
\begin{enumerate}
  \item[\textbf{RQ1}] Does continued pretraining on a learner corpus improve AES performance on English proficiency tests compared to general-domain transformer baselines?
  \item[\textbf{RQ2}] Does continued pretraining on a learner corpus improve transferability when transferring from one English proficiency test dataset to another under few-shot settings?
\end{enumerate}
\section{Background and Related Work}
Automated Essay Scoring on English proficiency tests has been the topic of research for over 20 years. Early AES studies focusing on this specific assessment setting often relied on classical machine learning techniques and feature engineering \cite{burstein-chodorow-1999-automated, lonsdale-strong-krause-2003-automated, yannakoudakis-etal-2011-new}. While these early studies demonstrated that AES could be effectively applied to learner writing, their approaches were typically developed and evaluated within relatively fixed assessment settings, where the prompts for essays are static.

Since the emergence of the Transformer architecture, AES research has shifted toward language models. In the context of English proficiency assessment, several studies have explored models such as BERT and DeBERTa for scoring learner essays \cite{qiu-etal-2024-large-language, schmalz-brutti-2021-automatic}. However, most of this work relies on general-domain transformers pretrained on corpora such as BookCorpus and Wikipedia, which consist largely of native-written text \cite{devlin-etal-2019-bert}. This creates a mismatch between the pretraining data and the target domain, since learner writing often contains grammatical errors, awkward phrasing, and unconventional lexical usage. As a result, general-domain language models may be less effective at capturing the characteristics of learner language, which can in turn limit AES performance and robustness.

A common approach to addressing distribution mismatch is domain adaptation. In AES, domain adaptation has been explored previously, but existing studies have focused largely on addressing the problem of prompt mismatch, where models suffer performance degradation when scoring essays written in response to unseen prompts \cite{phandi-etal-2015-flexible, cao2020domain, jiang-etal-2023-improving}. While prompt-level domain adaptation has proven effective in such settings, it is less suitable for AES in English proficiency tests, where there is no fixed prompt because the prompts are intentionally designed to vary across different topics and task types to effectively assess the language proficiency of the test taker. In this context, the distribution of learner language itself becomes a more appropriate target for adaptation.

A common domain adaptation approach widely used with Transformer-based models is domain-adaptive pretraining (DAPT) \cite{gururangan-etal-2020-dont}. In this method, a pretrained language model undergoes further pretraining on text drawn from the target domain. In the context of computer-assisted language learning, \citet{stearns-etal-2024-evaluating} applied DAPT to the BERT model using the EFCAMDAT corpus, a dataset consisting of learner writings from different levels of proficiency. Their findings demonstrate the promise of this approach, as the adapted model achieved strong results on intrinsic evaluations. However, the study did not examine whether these gains transfer to downstream tasks through task-specific fine-tuning.

Overall, prior AES work has largely relied on general-domain pretrained models or prompt-level domain adaptation. In English proficiency testing, however, prompts vary by design, making learner-language distribution a more stable adaptation target. We therefore examine whether DAPT on learner writing improves both in-domain scoring and cross-dataset robustness.

\section{Experimental Setup}
\subsection{Data}
\subsubsection{Pretraining Corpus}
\paragraph{EFCAMDAT.} EFCAMDAT is a large open-access corpus of L2 English learner writing,  spanning CEFR levels A1–C1 \cite{huang2017efcamdat_users,geertzen2014efcamdat}. The corpus comprises texts written by learners from diverse demographic backgrounds, representing 198 nationalities. In this study, we use the cleaned version released by \citet{efcamdat_cleaned}, which removes markup-heavy texts, duplicates, non-English content, ultra-short scripts, and extreme-length outliers, resulting in 45.39 million word tokens.
\subsubsection{Downstream datasets}
\paragraph{CLC FCE Dataset.} The CLC FCE dataset contains 1244 scripts written by upper-intermediate English as a Second or Other Language (ESOL) learners \cite{yannakoudakis-etal-2011-new}. Each script contains two responses to the two tasks in the writing section of the First Certificate in English (FCE) exam, which asks candidates to write a letter, a report, an article, a composition, or a short story between 200 and 400 words. In this dataset, each text is labeled with two different score annotations: an overall score that the script's author achieved across both writing tasks, and a score assigned to that specific response. In this study, we use the latter score as the target variable when fine-tuning and evaluating downstream performance.

\paragraph{IELTS Writing Scored Essays.} The IELTS Writing Scored Essays is a publicly available dataset on Kaggle\footnote{https://www.kaggle.com/datasets/mazlumi/ielts-writing-scored-essays-dataset}. This dataset was previously utilized by \citet{liew_2025_on} and \citet{qiu-etal-2024-large-language} in studies on the use of Large Language Models in AES, making it a good fit for our study. The dataset contains 1274 unique essays, covering both IELTS Writing Task 1 and Task 2, each annotated with a score ranging from 0 to 9. However, in this study, we only use IELTS Task 2 essays in the experiments. Task 1 essays were excluded because the dataset did not contain the figures or diagrams referenced in the prompts, making these responses incomplete for reliable modeling. As a result, the final dataset used in our experiments consists of 710 instances.

\subsection{Preprocessing}
For IELTS, we removed samples with missing texts or labels, duplicate entries, invalid scores, essays shorter than 100 characters, and removed non-English characters from the remaining essay text. For FCE, the response-level grades used here are categorical grade codes, which later AES work maps to a 0–20 numerical scale \cite{gaudeau-2025-beyond, zhang-etal-2018-effect}. In our study, we adopt the conversion scheme of \citet{zhang-etal-2018-effect} to map the original FCE grades to this 0–20 scale. 

\subsection{Domain-adaptive Pretraining (DAPT)} 
In this study, we perform domain-adaptive continued pretraining on three transformer-based models: BERT \cite{devlin-etal-2019-bert}, RoBERTa \cite{liu2019robertarobustlyoptimizedbert}, and DistilBERT \cite{sanh2020distilbertdistilledversionbert}, using the EFCAMDAT corpus as pretraining data. We selected the models above because they have been implemented in prior AES research \cite{wang-etal-2022-use, gui-2025-develop, qiu-etal-2024-large-language}, thus creating a well-established foundation for this study. BERT, in particular, has previously been used as the base model for continued pretraining on the EFCAMDAT corpus by \citet{stearns-etal-2024-evaluating}. Their study reported positive gains in intrinsic evaluation relative to its original checkpoint, which motivates us to examine whether such benefits also transfer to downstream AES on English proficiency assessment essays. For the specific pretraining details, see Appendix \ref{sec:training_details}.

Following \citet{stearns-etal-2024-evaluating}, we conduct continued pretraining with a Masked Language Modeling objective, masking 15\% of subword tokens in the training data and training the model to predict the masked items from their surrounding context. Their study also mentions filtering out texts with more than 512 tokens, which we also follow.
\subsection{Experiments} \label{sec:experiments}

\begin{figure*}[t]
    \centering
    \includegraphics[width=0.8\textwidth]{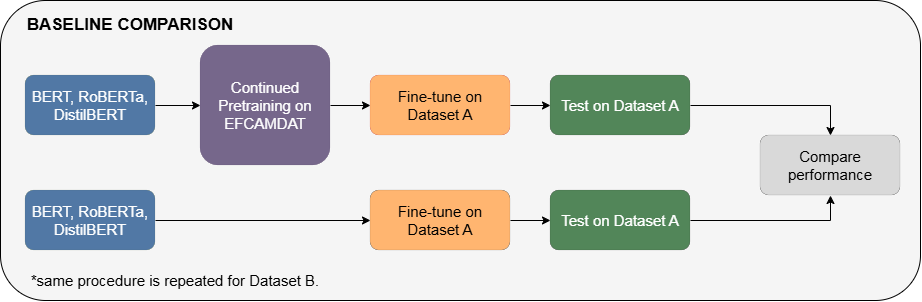}
    \caption{Baseline comparison setup. For each encoder architecture, the domain-adapted model is obtained by continuing pretraining from the corresponding base checkpoint on EFCAMDAT. The adapted and non-adapted models are then fine-tuned and evaluated on the same downstream dataset, and their performance is compared under identical fine-tuning settings. The same procedure is repeated for both downstream datasets.}
    \label{fig:Baseline}
\end{figure*}

\begin{figure*}[t]
    \centering
    \includegraphics[width=\textwidth]{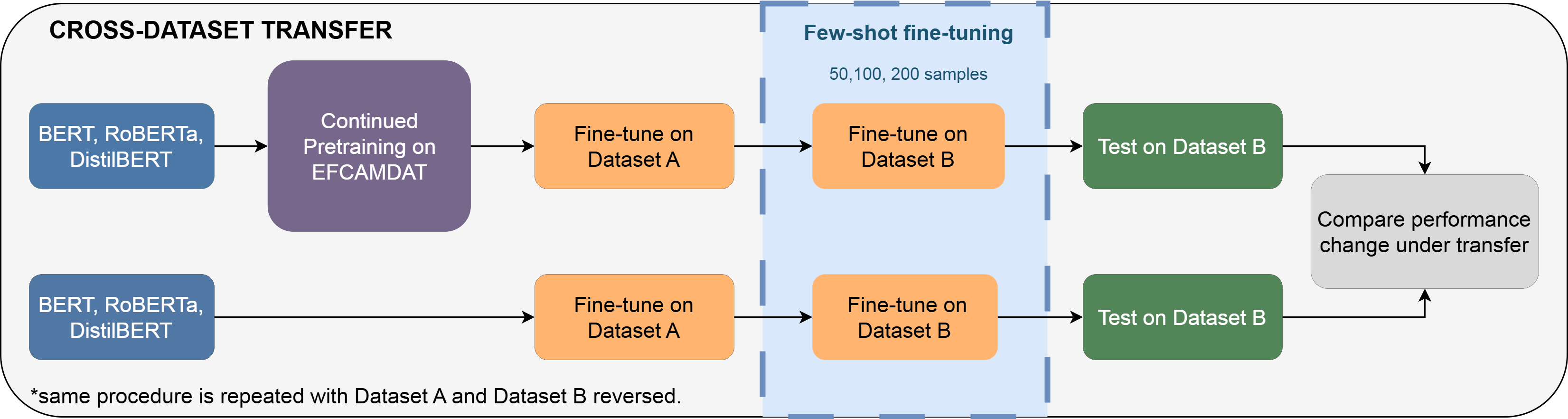}
    \caption{Cross-dataset transfer setup. For each encoder architecture, both the domain-adapted model and its corresponding non-adapted baseline are first fine-tuned on a source dataset, then adapted to the target dataset using a few-shot fine-tuning stage with 50, 100, and 200 target training samples, and finally evaluated on the target test set. Transfer-induced performance change is compared between each adapted model and its corresponding base checkpoint. The same procedure is repeated in both transfer directions.}
    \label{fig:cross-dataset}
\end{figure*}

To answer our research questions, we conduct two main experiments: baseline comparison and cross-dataset transfer. Figures \ref{fig:Baseline} and \ref{fig:cross-dataset} visualize these experiments.
\paragraph{Baseline Comparison.} In this experiment, the domain-adapted models are fine-tuned and evaluated separately on the FCE and IELTS datasets. Their performance is then compared with that of the corresponding non-adapted base models to determine whether DAPT on EFCAMDAT improves downstream scoring performance on the two datasets.

\paragraph{Cross-dataset Transfer.} In this experiment, we examine downstream cross-dataset transfer learning in both directions: from IELTS to FCE and vice versa. For each direction, the model is first fine-tuned on the source dataset, then subjected to an additional few-shot adaptation stage on the target dataset using 50, 100, and 200 training samples, before being evaluated on the target test set. Because the two datasets use different scoring ranges, the in-domain models are fine-tuned on normalized scores. We then compare the performance change associated with cross-dataset transfer between the domain-adapted models and their base checkpoints. This allows us to investigate whether continued pretraining on a learner corpus improves cross-dataset transferability, either by reducing performance degradation or by enabling larger gains under cross-dataset transfer.

\subsection{Evaluation Protocol and Metrics}
In our experiments, both downstream datasets are split with a 70/15/15 ratio for training, validating, and evaluation. For the FCE dataset in particular, we split by the provided author ID instead of individual text. This prevents essays written by the same learner from appearing in both training and evaluation sets, thereby reducing the risk of author-specific leakage and yielding a more reliable estimate of generalization. For all experiments, final performance is reported on the held-out test set that is not used during training or model selection. To ensure fairness, all experiments are conducted with the same hyperparameters and a fixed seed across runs. For the detailed hyperparameters used, see Appendix \ref{sec:training_details}. 

We conduct our experiments in a prompt-unaware setting, meaning that no explicit prompt information is provided to the model. This allows us to better isolate the effect of domain-adaptive pretraining on the linguistic characteristics of the essays themselves, rather than on prompt-specific cues. Since continued pretraining is intended to improve the model’s representation of learner writing, a prompt-unaware design provides a clearer test of whether such gains come from the essay text alone. This also provides a cleaner setting for our cross-dataset transfer experiment, where prompt mismatch can otherwise complicate the outcome.

Following prior AES work, we report RMSE, Quadratic Weighted Kappa (QWK) \cite{cohen1968weighted}, Pearson's correlation, and Spearman's correlation, because they are commonly reported metrics for this task \cite{phandi-etal-2015-flexible}. However, \citet{yannakoudakis-cummins-2015-evaluating} have highlighted limitations of both correlation-based measures and kappa-based agreement coefficients for automated text scoring evaluation. We therefore treat RMSE as the primary metric, which is also used by \citet{gaudeau-2025-beyond} as the main metric for validation and model selection.  Metrics are computed on the original score scales after denormalization. For QWK, continuous predictions are mapped to the nearest valid score category.

\section{Experimental Results}
In this section, we present and discuss the results obtained from the experiments described in Section \ref{sec:experiments}.
\subsection{Baseline Comparison}\label{sec:baseline}
Table~\ref{tab:baseline_results} presents the in-domain baseline comparison results. On IELTS Task 2, DAPT improves DistilBERT in terms of RMSE and QWK, but slightly lowers its correlation scores. In contrast, DAPT degrades both BERT and RoBERTa across all reported metrics, with the largest RMSE degradation observed for RoBERTa. On the FCE dataset, continued pretraining improves BERT in terms of RMSE and QWK, although its Spearman and Pearson correlations slightly decrease. For DistilBERT, DAPT leads to slightly weaker performance across all metrics, while for RoBERTa it improves QWK but worsens RMSE and correlation scores. These results suggest that continued pretraining on EFCAMDAT does not uniformly improve in-domain AES performance, and its effect differs across downstream datasets and the underlying encoder.
\begin{table*}[t]
\centering
{\small
\setlength{\tabcolsep}{4pt}

\begin{minipage}{0.48\textwidth}
\centering
\textbf{IELTS Task 2}

\vspace{0.3em}
\begin{tabular}{llrrrr}
\toprule
Model & Variant & RMSE & QWK & Sp. & Pe. \\
\midrule
BERT
& Non-adapted
& \textbf{0.632}
& \textbf{0.712}
& \textbf{0.784}
& 0.785 \\

& DAPT
& 0.737
& 0.689
& 0.764
& 0.778 \\

\addlinespace[0.15em]

RoBERTa
& Non-adapted
& 0.706
& 0.661
& 0.783
& \textbf{0.809} \\

& DAPT
& 0.852
& 0.489
& 0.739
& 0.766 \\

\addlinespace[0.15em]

DistilBERT
& Non-adapted
& 0.713
& 0.700
& 0.732
& 0.737 \\

& DAPT
& 0.691
& 0.705
& 0.725
& 0.730 \\

\bottomrule
\end{tabular}
\end{minipage}
\hfill
\begin{minipage}{0.48\textwidth}
\centering
\textbf{FCE}

\vspace{0.3em}
\begin{tabular}{llrrrr}
\toprule
Model & Variant & RMSE & QWK & Sp. & Pe. \\
\midrule
BERT
& Non-adapted
& 2.316
& 0.571
& 0.677
& 0.684 \\

& DAPT
& 2.101
& 0.598
& 0.652
& 0.678 \\

\addlinespace[0.15em]

RoBERTa
& Non-adapted
& 2.138
& 0.513
& \textbf{0.686}
& \textbf{0.695} \\

& DAPT
& 2.200
& 0.554
& 0.667
& 0.683 \\

\addlinespace[0.15em]

DistilBERT
& Non-adapted
& \textbf{2.080}
& \textbf{0.614}
& 0.667
& 0.675 \\

& DAPT
& 2.115
& 0.587
& 0.639
& 0.665 \\

\bottomrule
\end{tabular}
\end{minipage}
}

\caption{In-domain AES performance on IELTS Task 2 and FCE. Sp.=Spearman and Pe.=Pearson. DAPT denotes the corresponding encoder further pretrained on EFCAMDAT, while Non-adapted denotes the base checkpoint without further pretraining. Best performance achieved on each dataset is shown in bold.}
\label{tab:baseline_results}
\end{table*}

\subsection{Cross-dataset Transfer}\label{sec:cross-dataset_transfer_results}

\begin{table*}[t]
\centering
{\small
\setlength{\tabcolsep}{3pt}

\begin{minipage}{0.49\textwidth}
\centering
\textbf{IELTS $\rightarrow$ FCE}

\vspace{0.3em}
\begin{tabular}{llrrrr}
\toprule
Model & $n$ & RMSE & QWK & Sp. & Pe. \\
\midrule
BERT & 50  & -0.369 & -0.015 & -0.013 & -0.012 \\
     & 100 & -0.395 & -0.058 & \textbf{0.031} & \textbf{0.025} \\
     & 200 & -0.270 & -0.103 & \textbf{0.005} & -0.012 \\
\midrule
RoBERTa & 50  & -0.169 & -0.206 & -0.140 & -0.153 \\
        & 100 & \textbf{0.100} & -0.010 & \textbf{0.039} & \textbf{0.027} \\
        & 200 & \textbf{0.191} & -0.096 & -0.006 & -0.020 \\
\midrule
DistilBERT & 50  & \textbf{0.062} & \textbf{0.031} & -0.001 & \textbf{0.017} \\
           & 100 & \textbf{0.045} & \textbf{0.048} & \textbf{0.024} & \textbf{0.037} \\
           & 200 & \textbf{0.049} & \textbf{0.057} & \textbf{0.022} & \textbf{0.024} \\
\bottomrule
\end{tabular}
\end{minipage}
\hfill
\begin{minipage}{0.49\textwidth}
\centering
\textbf{FCE $\rightarrow$ IELTS}

\vspace{0.3em}
\begin{tabular}{llrrrr}
\toprule
Model & $n$ & RMSE & QWK & Sp. & Pe. \\
\midrule
BERT & 50  & -0.066 & -0.324 & -0.032 & -0.043 \\
     & 100 & \textbf{0.010} & -0.082 & -0.020 & -0.023 \\
     & 200 & -0.025 & -0.125 & -0.025 & -0.040 \\
\midrule
RoBERTa & 50  & \textbf{0.144} & \textbf{0.096} & \textbf{0.023} & -0.011 \\
        & 100 & \textbf{0.113} & -0.054 & -0.006 & -0.025 \\
        & 200 & \textbf{0.012} & \textbf{0.069} & \textbf{0.018} & \textbf{0.005} \\
\midrule
DistilBERT & 50  & -0.068 & -0.083 & -0.055 & -0.053 \\
           & 100 & -0.067 & -0.044 & -0.037 & -0.049 \\
           & 200 & -0.053 & -0.043 & -0.028 & -0.033 \\
\bottomrule
\end{tabular}
\end{minipage}
}

\caption{DAPT advantage in few-shot cross-test transfer. Each value compares the transfer-induced performance change of a DAPT model against its corresponding base model. Positive values indicate a more favorable transfer outcome for DAPT. RMSE is treated inversely. Sp.=Spearman; Pe.=Pearson.}
\label{tab:cross_transfer_dapt_advantage}
\end{table*}

Table~\ref{tab:cross_transfer_dapt_advantage} summarizes the relative advantage of DAPT in the cross-dataset transfer experiment. We compute the DAPT advantage in two steps:
\begin{enumerate}
    \item For each model and each value of $n$, we compute the transfer-induced performance change by comparing the $n$-shot cross-dataset transfer result with the corresponding in-domain result.
    \item We then compare this performance change between each DAPT model and its corresponding non-DAPT baseline.
\end{enumerate}
For example, if RMSE increases by 0.20 after transfer for the non-adapted model but by only 0.12 for its DAPT counterpart, the DAPT advantage is \(0.20-0.12=0.08\), indicating that DAPT incurs a smaller transfer-related performance loss. For QWK, Spearman, and Pearson, where higher values are better, larger DAPT advantage values indicate better performance preservation or improvement. For RMSE, where lower values are better, the sign is reversed so that positive values consistently indicate a more favorable transfer outcome. Thus, a positive DAPT advantage means that the DAPT model preserves performance better under transfer, either by suffering a smaller performance drop or by obtaining a larger transfer gain. The complete set of cross-transfer results used to compute these values is reported in Appendix~\ref{app:cross-dataset_transfer_results}.

The results are inconsistent across transfer directions, metrics, and models. In the IELTS to FCE transfer direction, DAPT BERT shows mostly negative advantages, suggesting that continued pretraining does not improve transfer robustness for BERT in this direction. DAPT RoBERTa produces mixed results, with positive RMSE advantages at 100 and 200 shots, but mostly negative QWK and correlation advantages. In contrast, DAPT DistilBERT shows the clearest positive pattern in this direction, with positive advantages for RMSE and QWK across all shot sizes and positive correlation advantages in most cases.

The reversed direction, with FCE as the source dataset, shows a different pattern. DAPT BERT again provides little evidence of transfer improvement, with negative advantages across most metrics and shot sizes. DAPT RoBERTa performs comparatively better in this direction, showing positive RMSE advantages across all shot sizes and positive advantages for several QWK and Spearman scores. However, this improvement is not uniform, as some correlation values remain negative. DAPT DistilBERT, which showed the strongest pattern in the IELTS to FCE direction, performs consistently worse than its base counterpart in the reverse direction, with negative advantages across all reported metrics and shot sizes. Overall, these results suggest that learner-domain DAPT does not consistently improve cross-test transferability, and its effect depends strongly on the transfer direction, model architecture, and evaluation metric.
\section{Discussion and Analysis}\label{sec:discussion}
We hypothesize that the mixed effects of DAPT on the full pretraining corpus are partly attributable to mismatches between the EFCAMDAT corpus and the downstream datasets, particularly in terms of proficiency profile, genre, and communicative purpose. This hypothesis is based on the known proficiency imbalance of EFCAMDAT, which is skewed toward lower-level A1--A2 learner writing \citep{stearns-etal-2024-evaluating}, as well as our preliminary qualitative inspection of corpus prompts and sample texts. To investigate this hypothesis, we conduct quantitative analyses between CEFR proficiency subsets of the EFCAMDAT and the downstream datasets, particularly in terms of vocabulary distribution and syntactic complexity. 
\subsection{Vocabulary Distribution} \label{sec:vocab}
To examine whether the pretraining corpus is lexically aligned with the downstream AES datasets, we compare vocabulary distributions across EFCAMDAT proficiency subsets and the downstream datasets. Specifically, we compute Jensen--Shannon divergence (JSD) \citep{Lin-JSD}, a symmetric divergence measure, between each CEFR-based EFCAMDAT subset and each downstream dataset. Following its use in NLP corpus comparison tasks \citep{lu-etal-2020-diverging}, we represent each corpus as a vocabulary probability distribution, where each token is assigned a probability based on its normalized frequency in that corpus.

\begin{equation}
\begin{aligned}
\mathrm{JSD}(P \,\|\, Q)
&= \frac{1}{2} D_{\mathrm{KL}}\!\left(P \,\middle\|\, \frac{P+Q}{2}\right) \\
&\quad + \frac{1}{2} D_{\mathrm{KL}}\!\left(Q \,\middle\|\, \frac{P+Q}{2}\right).
\end{aligned}
\label{eq:jsd}
\end{equation}

Here, \(P\) and \(Q\) denote the vocabulary probability distributions of two corpora, and \(D_{\mathrm{KL}}\) denotes Kullback--Leibler divergence \citep{kullback-leibler}. JSD measures how much each of the distributions diverges from their average distribution, \(\frac{P+Q}{2}\), with lower values indicating greater lexical similarity between the two corpora. 

The computed JSD results are reported in Table \ref{tab:jsd_efcamdat_downstream}. Overall, the values suggest that the FCE dataset is lexically closest to the B1 and B2 CEFR subsets of EFCAMDAT, indicating relatively stronger vocabulary distribution alignment with intermediate learner texts. In contrast, the IELTS dataset shows the lowest divergence from the C1 subset, suggesting closer lexical alignment with more advanced learner writing.

To provide a descriptive view of the lexical differences underlying the JSD results, we extract the 20 most frequent tokens in EFCAMDAT, IELTS, and FCE, both with and without stopwords (see Appendix~\ref{app:top_words} for the full list). When stopwords are included, EFCAMDAT and FCE show greater overlap in conversational high-frequency tokens, particularly pronouns such as \textit{i}, \textit{my}, and \textit{you}. By contrast, IELTS contains fewer personal pronouns among its most frequent tokens and includes more general argumentative tokens, such as \textit{people}, \textit{their}, and \textit{they}. The stopword-removed comparison further illustrates this pattern: EFCAMDAT and FCE contain more everyday and interactional lexical items, such as \textit{like}, \textit{good}, \textit{think}, and \textit{want}, whereas IELTS contains more topic-general and argumentative vocabulary, including \textit{society}, \textit{countries}, \textit{students}, \textit{however}, and \textit{example}. Overall, these patterns suggest a noticeable mismatch in genre and communicative purpose between EFCAMDAT and IELTS, whereas the mismatch between EFCAMDAT and FCE appears comparatively smaller.
\begin{table}[t]
\centering
\small
\begin{tabular}{lcc}
\toprule
CEFR subset & FCE & IELTS \\
\midrule
A1 & 0.3006 & 0.4322 \\
A2 & 0.1783 & 0.3398 \\
B1 & \textbf{0.1373} & 0.2235 \\
B2 & 0.1397 & 0.1810 \\
C1 & 0.1575 & \textbf{0.1574} \\
\bottomrule
\end{tabular}
\caption{Jensen--Shannon divergence between each EFCAMDAT CEFR proficiency subset and each downstream AES dataset. Lower values indicate greater vocabulary distribution similarity. The lowest JSD for each downstream dataset is shown in bold.}
\label{tab:jsd_efcamdat_downstream}
\end{table}

\subsection{Syntactic Complexity}\label{sec:syntactic}
\begin{table*}[t]
\centering
\small
\begin{tabular}{lrrrrrrr}
\toprule
Source & MLT & MLC & C/T & DC/C & CT/T & CP/T & CN/T \\
\midrule
EFCAMDAT A1 & 9.006 & 8.259 & 1.115 & 0.070 & 0.087 & 0.285 & 0.626 \\
EFCAMDAT A2 & 10.334 & 8.469 & 1.227 & 0.142 & 0.183 & 0.265 & 0.745 \\
EFCAMDAT B1 & 13.655 & 9.649 & 1.434 & 0.249 & 0.319 & 0.302 & 1.227 \\
EFCAMDAT B2 & 14.408 & 9.758 & 1.537 & 0.297 & 0.384 & 0.326 & 1.412 \\
EFCAMDAT C1 & 15.827 & 10.554 & 1.547 & 0.307 & 0.400 & 0.374 & 1.692 \\
FCE train    & 14.227 & 8.299 & 1.729 & 0.351 & 0.467 & 0.222 & 1.208 \\
IELTS train  & 20.726 & 11.036 & 1.915 & 0.412 & 0.543 & 0.552 & 2.620 \\
\bottomrule
\end{tabular}

\caption{Syntactic complexity profiles of EFCAMDAT proficiency subsets and downstream training datasets.}
\label{tab:syntactic_complexity_profiles}

\vspace{0.3em}
\begin{minipage}{0.95\textwidth}
\footnotesize
\textit{Note.} MLT = mean length of T-unit; MLC = mean length of clause; C/T = clauses per T-unit; DC/C = dependent clauses per clause; CT/T = complex T-units per T-unit; CP/T = coordinate phrases per T-unit; CN/T = complex nominals per T-unit. A T-unit refers to an independent clause together with any dependent clauses attached to it.
\end{minipage}
\end{table*}

To examine syntactic alignment between the pretraining corpus and the downstream datasets, we use NeoSCA\footnote{https://github.com/tanloong/neosca}, a modern implementation of the L2 Syntactic Complexity Analyzer (L2SCA) \cite{l2sca} to extract syntactic complexity features from each dataset. NeoSCA is applied at the text level, and the resulting feature values are then aggregated to obtain corpus-level syntactic profiles. For the downstream datasets, we apply NeoSCA to all available texts. For the EFCAMDAT proficiency subsets, however, applying NeoSCA to the full corpus would be computationally expensive due to the large size of the dataset. Therefore, we randomly sample 500 texts from each CEFR proficiency subset and apply the same extraction procedure to each sample. Although NeoSCA outputs 14 syntactic complexity indices, we report a focused shortlist of seven features most relevant to our mismatch hypothesis, which capture differences in production-unit length, clausal density, subordination, coordination, and nominal/phrasal complexity. The results are available in Table ~\ref{tab:syntactic_complexity_profiles}.

Overall, the syntactic patterns are broadly consistent with the earlier vocabulary distribution results in Section \ref{sec:vocab}. Among the EFCAMDAT proficiency subsets, FCE appears most closely aligned with the B1 and B2 subsets, suggesting that its syntactic profile is closer to intermediate-level learner writing. IELTS, by contrast, is closest to the C1 subset, reflecting its generally higher syntactic complexity. However, the gap between IELTS and even the closest EFCAMDAT subset remains substantial, indicating that IELTS is still relatively misaligned with the available EFCAMDAT proficiency levels. This supports the view that dataset mismatch is not only lexical, but also reflected in syntactic complexity.

\section{Ablation Study}\label{sec:ablation_study}
In this section, we conduct a proficiency-based ablation study, in which continued pretraining is performed on selected CEFR proficiency subsets of EFCAMDAT rather than on the full corpus. This allows us to examine whether downstream AES performance is sensitive to the proficiency composition of the DAPT corpus and, in particular, whether adaptation on learner writing that is more closely aligned with the downstream dataset can outperform full-corpus DAPT. To assess whether this effect generalizes across encoder architectures, the ablation is conducted using BERT, RoBERTa, and DistilBERT. Because the number of C1-level texts in EFCAMDAT is limited, we combine B2 and C1 texts to form the highest-proficiency subset. This grouping provides a more viable amount of pretraining data while still representing the upper-proficiency range of the learner corpus.
The available A1--A2, B1--B2, and B2--C1 subsets contain approximately 35.9M, 20.1M, and 8.6M tokens, respectively. To control for this substantial difference in data quantity, we cap each proficiency-specific DAPT condition at 8 million tokens, corresponding approximately to the largest common pretraining budget that can be supported by all three subsets. This allows the ablation to more directly examine the effect of proficiency composition rather than differences in the amount of pretraining data.
Table \ref{tab:ablation_results} presents the in-domain scoring results of the proficiency-based ablation. Overall, the effect of proficiency composition varies across model architectures and downstream datasets, although the BERT results show the clearest correspondence with the lexical and syntactic alignment analyses reported in Section \ref{sec:discussion}. 
\begin{table*}[t]
\centering
\small
\setlength{\tabcolsep}{4pt}
\begin{tabular}{lcccccccc}
\toprule
& \multicolumn{4}{c}{\textbf{FCE}} 
& \multicolumn{4}{c}{\textbf{IELTS}} \\
\cmidrule(lr){2-5}
\cmidrule(lr){6-9}
\textbf{Pretraining Data} 
& \textbf{RMSE} & \textbf{QWK} & \textbf{Sp.} & \textbf{Pe.}
& \textbf{RMSE} & \textbf{QWK} & \textbf{Sp.} & \textbf{Pe.} \\
\midrule

\multicolumn{9}{l}{\textbf{BERT}} \\
No EFCAMDAT   
& 2.316 & 0.571 & \textbf{0.677} & \textbf{0.684}
& \textbf{0.632} & \textbf{0.712} & 0.784 & 0.785 \\

Full EFCAMDAT 
& 2.101 & 0.598 & 0.652 & 0.678
& 0.737 & 0.689 & 0.764 & 0.778 \\

A1--A2 only   
& 2.177 & 0.581 & 0.647 & 0.665
& 0.735 & 0.687 & \textbf{0.795} & \textbf{0.793} \\

B1--B2 only   
& \textbf{2.064} & \textbf{0.609} & 0.662 & 0.679
& 0.758 & 0.663 & 0.772 & 0.777 \\

B2--C1 only   
& 2.143 & 0.550 & 0.636 & 0.653
& 0.695 & 0.627 & 0.725 & 0.751 \\

\midrule
\multicolumn{9}{l}{\textbf{RoBERTa}} \\
No EFCAMDAT   
& \textbf{2.138} & 0.513 & 0.686 & 0.695
& \textbf{0.706} & 0.661 & 0.783 & 0.809 \\

Full EFCAMDAT 
& 2.200 & 0.554 & 0.667 & 0.683
& 0.852 & 0.489 & 0.739 & 0.766 \\

A1--A2 only   
& 2.414 & 0.388 & 0.626 & 0.631
& 0.715 & 0.655 & \textbf{0.793} & \textbf{0.816} \\

B1--B2 only   
& 2.173 & \textbf{0.568} & 0.650 & 0.666
& 0.714 & \textbf{0.669} & 0.784 & 0.806 \\

B2--C1 only   
& 2.206 & 0.475 & \textbf{0.692} & \textbf{0.706}
& 0.786 & 0.620 & 0.776 & 0.802 \\

\midrule
\multicolumn{9}{l}{\textbf{DistilBERT}} \\
No EFCAMDAT   
& \textbf{2.080} & \textbf{0.614} & \textbf{0.667} & \textbf{0.675}
& 0.713 & 0.700 & \textbf{0.732} & \textbf{0.737} \\

Full EFCAMDAT 
& 2.115 & 0.587 & 0.639 & 0.665
& \textbf{0.691} & \textbf{0.705} & 0.725 & 0.730 \\

A1--A2 only   
& 2.126 & 0.563 & 0.636 & 0.658
& 0.718 & 0.702 & 0.718 & 0.726 \\

B1--B2 only   
& 2.121 & 0.575 & 0.648 & 0.664
& 0.757 & 0.668 & 0.706 & 0.719 \\

B2--C1 only   
& 2.135 & 0.577 & 0.654 & 0.667
& 0.770 & 0.676 & 0.707 & 0.711 \\

\bottomrule
\end{tabular}

\caption{Ablation results across model architectures on FCE and IELTS.``No EFCAMDAT'' denotes the corresponding pretrained encoder without
learner-domain DAPT. Sp.=Spearman and Pe.=Pearson. The best result for
each model, dataset, and metric is shown in bold.}
\label{tab:ablation_results}
\end{table*}

For BERT, DAPT on the B1--B2 subset achieves the lowest RMSE and highest QWK among all DAPT variants on FCE, while achieving comparable but slightly lower Spearman and Pearson correlations than the non-adapted baseline. This is partially consistent with the earlier observation that FCE is most closely aligned with the B1--B2 EFCAMDAT subset in terms of vocabulary distribution and syntactic complexity. On the IELTS dataset, the best-performing DAPT variant in terms of RMSE is the B2–C1 subset. However, this variant still does not outperform the non-adapted BERT checkpoint, which remains strongest across RMSE and QWK. This result aligns with earlier analysis in Section \ref{sec:syntactic}, where the C1 EFCAMDAT texts were found to be closest to IELTS in terms of syntactic complexity, yet still different by a substantial margin. Meanwhile, the A1--A2 subset achieves the highest Spearman and Pearson correlations, although the differences from the non-adapted model are relatively small.

Similar results are observed with RoBERTa. On the FCE dataset, the B1--B2 subset is the best-performing DAPT variant in terms of RMSE, although it does not outperform the non-adapted baseline, whose RMSE is slightly lower by 0.035. Nevertheless, the B1--B2 variant achieves the highest QWK among all RoBERTa variants, while the B2--C1 subset yields the strongest Spearman and Pearson correlations. Since B1--B2 and B2--C1 overlap, it is unsurprising that neither consistently dominates across metrics. This is also consistent with Section \ref{sec:vocab}, where Table \ref{tab:jsd_efcamdat_downstream} shows relatively similar JSD values for B1, B2, and C1 with respect to FCE. A broadly similar pattern is observed on the IELTS dataset. The B1--B2 subset achieves the lowest RMSE among the DAPT variants, although its RMSE remains marginally higher than that of the non-adapted baseline by 0.008. In contrast, the A1--A2 subset achieves the highest Spearman and Pearson correlations, a pattern also observed previously in the BERT experiments.

However, the above pattern does not carry over consistently to DistilBERT, which appears substantially less sensitive to the composition of the DAPT corpus. On FCE, all four DAPT variants remain close in performance across the evaluation metrics, although each is outperformed by the non-adapted baseline. On IELTS, however, full-corpus EFCAMDAT DAPT achieves the lowest RMSE and highest QWK, while the non-adapted baseline remains strongest in terms of Spearman and Pearson correlations. 

In addition to in-domain scoring, we further extend the proficiency-based ablation to the cross-dataset transfer experiment. However, due to the scope of this study, this extension is only conducted on the BERT model. Appendix \ref{app:ablated_transfer} presents the results for this experiment. The transfer results remain inconsistent across transfer directions, metrics, and pretraining subsets. In other words, restricting DAPT to proficiency-aligned EFCAMDAT subsets does not consistently reduce the transfer gap between FCE and IELTS. This suggests that proficiency alignment alone is not sufficient to improve cross-dataset transferability, especially when the two downstream datasets differ in scoring scale, task design, genre, and communicative purpose. 

Overall, the ablation provides evidence that proficiency-specific DAPT can be preferable to full-corpus adaptation. Across the 24 individual metric comparisons presented in Table \ref{tab:ablation_results}, full-EFCAMDAT DAPT is outperformed by a proficiency-specific DAPT variant in 17 cases. Moreover, proficiency subsets that are more closely aligned with the downstream data can, in some settings, outperform the non-adapted baseline, as most clearly observed for BERT on FCE and partially for RoBERTa on the same dataset. However, these advantages are not consistent across encoder architectures, downstream datasets, or evaluation metrics. Taken together, the results suggest that proficiency composition is an important factor in learner-domain DAPT, but its effect is conditional rather than universally beneficial.

\section{Conclusion}
This study set out to examine whether learner-domain DAPT provides a reliable benefit for automated essay scoring on English proficiency tests. Rather than finding a uniform improvement from continued pretraining on EFCAMDAT, our results show a more conditional pattern. Full-corpus DAPT produced mixed effects across datasets, models, and metrics, suggesting that exposure to a broad learner corpus alone is not sufficient to guarantee better AES performance. Further lexical and syntactic analyses, followed by proficiency-based ablations, indicate that DAPT can improve in-domain essay scoring over the non-adapted baseline in some settings, with the clearest gains observed when the pretraining data are more closely aligned with the downstream assessment dataset. More importantly, these proficiency-specific DAPT variants also frequently outperform full-corpus EFCAMDAT adaptation, suggesting that the proficiency composition of the learner pretraining data plays an important role in determining the effectiveness of DAPT. However, these gains are not consistent across evaluation metrics, downstream datasets, or encoder architectures. Moreover, they do not reliably extend to the cross-dataset transfer setting, where DAPT does not consistently improve transferability across assessment contexts.

These findings suggest that corpus alignment should be considered when selecting data for DAPT on learner writing. Comparing proficiency distribution, genre, communicative purpose, lexical similarity, and syntactic complexity may help identify pretraining subsets that are better aligned with the target data and could yield better downstream AES performance. However, improvement is not guaranteed. Such alignment should therefore be treated as a useful screening criterion rather than a reliable predictor of DAPT effectiveness.

\section*{Limitations}
This study has several limitations. First, continued pretraining was conducted using only one learner corpus, EFCAMDAT. Although EFCAMDAT is large and well suited to learner-language adaptation, its proficiency distribution, task types, and communicative purposes may not represent learner writing more broadly. As our lexical and syntactic analyses show, the corpus is not equally aligned with the downstream assessment datasets, which limits the generalizability of the observed DAPT effects.

Second, the downstream evaluation is limited to two English proficiency test datasets, FCE and IELTS Task 2. The IELTS subset is relatively small after preprocessing, and the findings may therefore be sensitive to dataset composition and splits. Future work should evaluate learner-domain DAPT on additional proficiency tests and larger writing datasets.

Third, although the proficiency-based in-domain ablation is conducted across BERT, RoBERTa, and DistilBERT, the corresponding cross-dataset transfer ablation is limited to BERT due to the scope of the study and high computational cost. Therefore, while the in-domain results indicate that the effect of proficiency-focused DAPT varies across encoder architectures, further work is needed to determine whether similar architecture-specific patterns also hold for cross-dataset transfer.

Finally, experiments were conducted using a single fixed random seed, so small performance differences should be interpreted cautiously. Future work should assess variability across multiple runs.

\section*{Ethical Considerations}
While we plan to make the code used in this study publicly available, access to EFCAMDAT is governed by a License Agreement and is restricted to academic, non-commercial research, with conditions protecting copyright. For this reason, we will not redistribute the corpus or any text-derived data artifacts with our code, including the model checkpoints, as it risks the regulated terms being violated beyond our control. Researchers wishing to reproduce our experiments must obtain access to EFCAMDAT through the official provider. 
At the same time, we believe that making the trained models available would benefit future research by reducing computational cost and enabling more direct comparison with follow-up studies. We therefore have contacted the EFCAMDAT providers regarding the possibility of releasing the trained model checkpoints through the official EFCAMDAT website or another provider-approved access channel. Any such release would be subject to the providers’ approval and to the same access, copyright, and non-commercial research conditions governing the corpus.

\section*{Acknowledgements}
We would like to thank the curators of the EF Cambridge Open Language Database (EFCAMDAT) for making the learner corpus available to the research community.

The author also acknowledges the use of AI tools, including ChatGPT, for language polishing, code debugging, and discussion of manuscript structure. All research design decisions, experiments, analyses, interpretations, and final manuscript content were reviewed and controlled by the author.
\bibliography{custom}
\appendix
\section{Training Details}\label{sec:training_details}
In this appendix, we provide additional training details to support the reproducibility of our experiments. Specifically, Table \ref{tab:training_hyperparameters} reports the key hyperparameters used during domain-adaptive continued pretraining and downstream experiments.

\begin{figure*}[t]
    \centering
    \includegraphics[width=0.8\textwidth]{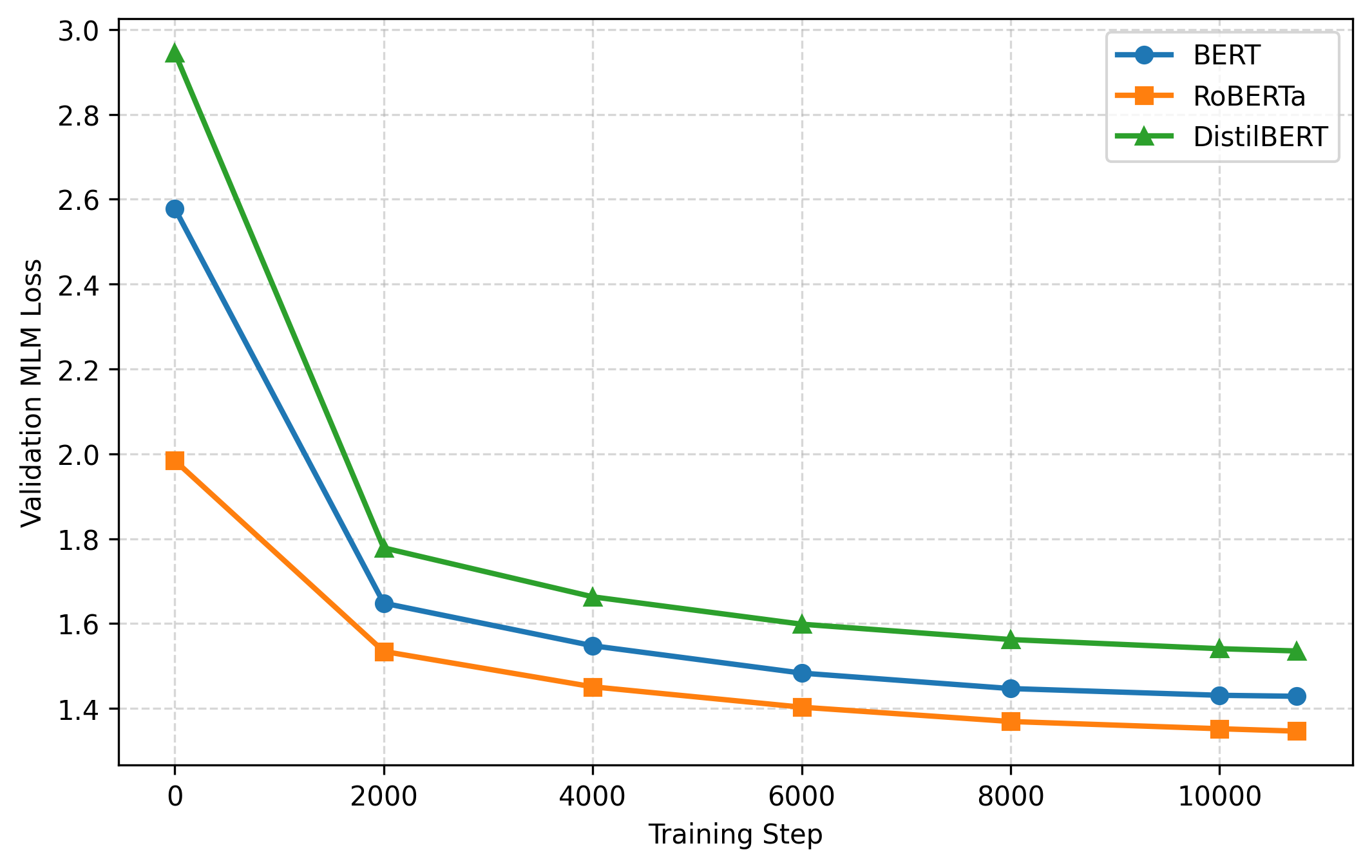}
    \caption{Validation MLM loss of BERT, RoBERTa, and DistilBERT during continued pretraining on EFCAMDAT. Lower values indicate better prediction of masked subword tokens on the held-out validation split. At step 0, the reported MLM loss is that of the corresponding base checkpoint without DAPT.}
    \label{fig:mlm_loss_curve}
\end{figure*}

Additionally, we report the validation MLM loss curves observed during continued pretraining to show that the adapted models learned from EFCAMDAT and that training proceeded without evidence of validation-loss divergence. The decision to train for only one epoch is supported by these curves: for all models, validation MLM loss decreases substantially early in training and begins to plateau toward the end of the epoch, suggesting that further training would likely yield diminishing returns. Figure~\ref{fig:mlm_loss_curve} visualizes these learning curves.

The code for reproducing the experiments in this study is available at \url{https://github.com/KatoTheFluffyWolf/DAPT-EFCAMDAT}. However, the repository does not contain the original or processed EFCAMDAT data, for the reasons discussed in Ethical Considerations.

\begin{table*}[t]
\centering
\small
\setlength{\tabcolsep}{6pt}
\begin{tabular}{cll}
\toprule
Stage & Hyperparameter / Setting & Value \\
\midrule
\multirow{12}{*}{Continued pretraining}
& BERT checkpoint & google-bert/bert-base-uncased \\
& RoBERTa checkpoint & FacebookAI/roberta-base \\
& DistilBERT checkpoint & distilbert/distilbert-base-uncased \\
& Total training steps & 10737 \\
& Hardware & NVIDIA L4 GPU \\
& Random seed & 42 \\
& Maximum sequence length & 512 \\
& MLM masking probability & 0.15 \\
& Training epochs & 1 \\
& Batch configuration & Train BS = 4, eval BS = 8, grad. accum. = 16, effective train BS = 64 \\
& Learning rate & $5 \times 10^{-5}$ \\
& Optimizer & AdamW \\
& Weight decay & 0.01 \\
& Warm-up & 6\% of total training steps \\
& Learning rate scheduler & Linear \\
& Precision & FP16 \\
& Validation split & 5\% \\
\midrule
\multirow{5}{*}{Downstream}
& Maximum sequence length & 512 \\
& Learning rate & $2 \times 10^{-5}$ \\
& Batch configuration & Train BS = 8, eval BS = 16 \\
& Training epochs & 5 \\
& Model selection metric & RMSE \\
& Score normalization & Min-max \\
& Early Stopping & No early stopping; checkpoint with best validation RMSE retained. \\
& Few-shot sampling & Stratified sampling by target score (seed = 42) \\
\bottomrule
\end{tabular}
\caption{Main hyperparameters used for continued pretraining and downstream experiments.}
\label{tab:training_hyperparameters}
\end{table*}

\section{Supplementary Experimental Result}

\subsection{Cross-dataset Transfer Full Results} \label{app:cross-dataset_transfer_results}
Table~\ref{tab:cross_transfer_full_results} reports the complete results of the cross-dataset transfer experiment described in Section~\ref{sec:experiments}. These results are used to compute the DAPT advantage values summarized in Table~\ref{tab:cross_transfer_dapt_advantage}.

\begin{table*}[t]
\centering
\scriptsize
\setlength{\tabcolsep}{4pt}
\renewcommand{\arraystretch}{1.08}
\begin{threeparttable}
\caption{Full cross-test transfer results. Values in parentheses show the difference from the corresponding model's in-domain result.}
\label{tab:cross_transfer_full_results}
\begin{tabular}{llrrrr}
\toprule
Model & $n$ & RMSE $\downarrow$ & QWK $\uparrow$ & Spearman $\uparrow$ & Pearson $\uparrow$ \\
\midrule

\multicolumn{6}{l}{\textbf{IELTS $\rightarrow$ FCE}} \\
\midrule
DAPT BERT & In-domain & 2.101 & 0.598 & 0.652 & 0.678 \\
          & 50  & 2.557 (+0.456) & 0.429 (-0.169) & 0.490 (-0.162) & 0.515 (-0.163) \\
          & 100 & 2.433 (+0.332) & 0.483 (-0.115) & 0.596 (-0.056) & 0.616 (-0.062) \\
          & 200 & 2.271 (+0.170) & 0.510 (-0.088) & 0.590 (-0.062) & 0.605 (-0.073) \\
\addlinespace

BERT      & In-domain & 2.316 & 0.571 & 0.677 & 0.684 \\
          & 50  & 2.403 (+0.087) & 0.417 (-0.154) & 0.528 (-0.149) & 0.533 (-0.151) \\
          & 100 & 2.253 (-0.063) & 0.514 (-0.057) & 0.590 (-0.087) & 0.597 (-0.087) \\
          & 200 & 2.216 (-0.100) & 0.586 (+0.015) & 0.610 (-0.067) & 0.623 (-0.061) \\
\addlinespace

DAPT RoBERTa & In-domain & 2.200 & 0.554 & 0.667 & 0.683 \\
             & 50  & 2.671 (+0.471) & 0.265 (-0.289) & 0.485 (-0.182) & 0.487 (-0.196) \\
             & 100 & 2.233 (+0.033) & 0.517 (-0.037) & 0.661 (-0.006) & 0.664 (-0.019) \\
             & 200 & 2.196 (-0.004) & 0.495 (-0.059) & 0.637 (-0.030) & 0.640 (-0.043) \\
\addlinespace

RoBERTa      & In-domain & 2.138 & 0.513 & 0.686 & 0.695 \\
             & 50  & 2.440 (+0.302) & 0.430 (-0.083) & 0.644 (-0.042) & 0.652 (-0.043) \\
             & 100 & 2.271 (+0.133) & 0.486 (-0.027) & 0.641 (-0.045) & 0.649 (-0.046) \\
             & 200 & 2.325 (+0.187) & 0.550 (+0.037) & 0.662 (-0.024) & 0.672 (-0.023) \\
\addlinespace

DAPT DistilBERT & In-domain & 2.115 & 0.587 & 0.639 & 0.665 \\
                & 50  & 2.395 (+0.280) & 0.432 (-0.155) & 0.493 (-0.146) & 0.525 (-0.140) \\
                & 100 & 2.297 (+0.182) & 0.530 (-0.057) & 0.570 (-0.069) & 0.602 (-0.063) \\
                & 200 & 2.208 (+0.093) & 0.589 (+0.002) & 0.590 (-0.049) & 0.626 (-0.039) \\
\addlinespace

DistilBERT      & In-domain & 2.080 & 0.614 & 0.667 & 0.675 \\
                & 50  & 2.422 (+0.342) & 0.428 (-0.186) & 0.522 (-0.145) & 0.518 (-0.157) \\
                & 100 & 2.307 (+0.227) & 0.509 (-0.105) & 0.574 (-0.093) & 0.575 (-0.100) \\
                & 200 & 2.222 (+0.142) & 0.559 (-0.055) & 0.596 (-0.071) & 0.612 (-0.063) \\

\midrule
\multicolumn{6}{l}{\textbf{FCE $\rightarrow$ IELTS}} \\
\midrule

DAPT BERT & In-domain & 0.737 & 0.689 & 0.764 & 0.778 \\
          & 50  & 0.832 (+0.095) & 0.375 (-0.314) & 0.699 (-0.065) & 0.699 (-0.079) \\
          & 100 & 0.743 (+0.006) & 0.628 (-0.061) & 0.727 (-0.037) & 0.742 (-0.036) \\
          & 200 & 0.737 (+0.000) & 0.660 (-0.029) & 0.748 (-0.016) & 0.755 (-0.023) \\
\addlinespace

BERT      & In-domain & 0.632 & 0.712 & 0.784 & 0.785 \\
          & 50  & 0.661 (+0.029) & 0.722 (+0.010) & 0.751 (-0.033) & 0.749 (-0.036) \\
          & 100 & 0.648 (+0.016) & 0.733 (+0.021) & 0.767 (-0.017) & 0.772 (-0.013) \\
          & 200 & 0.607 (-0.025) & 0.808 (+0.096) & 0.793 (+0.009) & 0.802 (+0.017) \\
\addlinespace

DAPT-RoBERTa & In-domain & 0.852 & 0.489 & 0.739 & 0.766 \\
             & 50  & 0.717 (-0.135) & 0.600 (+0.111) & 0.754 (+0.015) & 0.734 (-0.032) \\
             & 100 & 0.792 (-0.060) & 0.477 (-0.012) & 0.723 (-0.016) & 0.733 (-0.033) \\
             & 200 & 0.726 (-0.126) & 0.661 (+0.172) & 0.760 (+0.021) & 0.769 (+0.003) \\
\addlinespace

RoBERTa      & In-domain & 0.706 & 0.661 & 0.783 & 0.809 \\
             & 50  & 0.715 (+0.009) & 0.676 (+0.015) & 0.775 (-0.008) & 0.788 (-0.021) \\
             & 100 & 0.759 (+0.053) & 0.703 (+0.042) & 0.773 (-0.010) & 0.801 (-0.008) \\
             & 200 & 0.592 (-0.114) & 0.764 (+0.103) & 0.786 (+0.003) & 0.807 (-0.002) \\
\addlinespace

DAPT-DistilBERT & In-domain & 0.691 & 0.705 & 0.725 & 0.730 \\
                & 50  & 0.822 (+0.131) & 0.547 (-0.158) & 0.666 (-0.059) & 0.669 (-0.061) \\
                & 100 & 0.726 (+0.035) & 0.623 (-0.082) & 0.707 (-0.018) & 0.693 (-0.037) \\
                & 200 & 0.687 (-0.004) & 0.669 (-0.036) & 0.716 (-0.009) & 0.725 (-0.005) \\
\addlinespace

DistilBERT      & In-domain & 0.713 & 0.700 & 0.732 & 0.737 \\
                & 50  & 0.776 (+0.063) & 0.625 (-0.075) & 0.728 (-0.004) & 0.729 (-0.008) \\
                & 100 & 0.681 (-0.032) & 0.662 (-0.038) & 0.751 (+0.019) & 0.749 (+0.012) \\
                & 200 & 0.656 (-0.057) & 0.707 (+0.007) & 0.751 (+0.019) & 0.765 (+0.028) \\

\bottomrule
\end{tabular}
\begin{tablenotes}
\small
\item For RMSE, lower is better; therefore, a positive difference from in-domain means performance worsened, while a negative difference means performance improved. For QWK, Spearman, and Pearson, higher is better.
\end{tablenotes}
\end{threeparttable}
\end{table*}

\subsection{Top-k words}\label{app:top_words}
Table~\ref{tab:top20_tokens} reports the 20 most frequent tokens in EFCAMDAT, IELTS, and FCE, both before and after stopword removal. These token lists provide a descriptive view of the vocabulary patterns discussed in Section~\ref{sec:vocab}.
\begin{table*}[t]
\centering
\small
\setlength{\tabcolsep}{6pt}
\begin{tabular}{rlrlrlr}
\toprule
Rank 
& \multicolumn{2}{c}{EFCAMDAT} 
& \multicolumn{2}{c}{IELTS} 
& \multicolumn{2}{c}{FCE} \\
\cmidrule(lr){2-3} \cmidrule(lr){4-5} \cmidrule(lr){6-7}
& Token & Count & Token & Count & Token & Count \\
\midrule
\multicolumn{7}{l}{\textit{Panel A: All tokens}} \\
\midrule
1  & the   & 1,915,133 & the    & 7,742 & the   & 14,921 \\
2  & i     & 1,840,818 & to     & 5,825 & i     & 13,789 \\
3  & and   & 1,518,481 & of     & 4,445 & to    & 12,429 \\
4  & a     & 1,160,866 & and    & 4,150 & and   & 8,029 \\
5  & to    & 1,141,105 & in     & 3,665 & a     & 5,902 \\
6  & in    & 986,490   & a      & 3,236 & in    & 5,739 \\
7  & my    & 918,668   & is     & 2,773 & you   & 5,691 \\
8  & is    & 915,013   & that   & 2,560 & of    & 5,320 \\
9  & you   & 588,095   & for    & 1,986 & it    & 4,780 \\
10 & of    & 557,787   & are    & 1,681 & was   & 4,086 \\
11 & are   & 404,345   & it     & 1,672 & that  & 4,023 \\
12 & have  & 386,269   & people & 1,672 & is    & 3,899 \\
13 & for   & 384,402   & be     & 1,615 & for   & 3,638 \\
14 & at    & 341,735   & their  & 1,452 & my    & 3,196 \\
15 & i'm   & 312,911   & this   & 1,407 & have  & 3,163 \\
16 & it    & 298,756   & as     & 1,334 & we    & 2,643 \\
17 & like  & 285,750   & they   & 1,215 & be    & 2,408 \\
18 & very  & 284,760   & can    & 1,116 & at    & 2,377 \\
19 & that  & 279,017   & not    & 1,037 & would & 2,140 \\
20 & with  & 275,151   & have   & 1,003 & but   & 2,091 \\
\midrule
\multicolumn{7}{l}{\textit{Panel B: Non-stopword tokens}} \\
\midrule
1  & like   & 285,750 & people   & 1,672 & would    & 2,140 \\
2  & work   & 159,293 & many     & 529   & like     & 1,962 \\
3  & people & 158,429 & also     & 472   & show     & 1,388 \\
4  & good   & 142,338 & one      & 443   & time     & 1,159 \\
5  & years  & 125,333 & would    & 426   & people   & 1,095 \\
6  & one    & 109,100 & children & 413   & dear     & 974 \\
7  & day    & 108,973 & life     & 396   & think    & 964 \\
8  & live   & 108,392 & time     & 396   & good     & 940 \\
9  & job    & 103,874 & could    & 358   & could    & 869 \\
10 & lot    & 97,251  & believe  & 354   & money    & 858 \\
11 & time   & 97,084  & students & 342   & one      & 838 \\
12 & old    & 92,461  & however  & 341   & know     & 836 \\
13 & name   & 91,261  & countries& 330   & shopping & 766 \\
14 & city   & 91,138  & society  & 301   & life     & 765 \\
15 & going  & 88,269  & example  & 296   & first    & 756 \\
16 & many   & 85,801  & work     & 284   & also     & 735 \\
17 & think  & 84,975  & may      & 283   & clothes  & 658 \\
18 & get    & 84,087  & world    & 282   & really   & 618 \\
19 & two    & 83,406  & money    & 263   & much     & 612 \\
20 & love   & 82,810  & country  & 256   & want     & 591 \\
\bottomrule
\end{tabular}
\caption{Top 20 most frequent tokens in EFCAMDAT, IELTS, and FCE, reported with and without stopwords.}
\label{tab:top20_tokens}
\end{table*}

\subsection{Ablated Cross-dataset Transfer Results}
\label{app:ablated_transfer}
In this appendix, we report the results of the ablated cross-dataset transfer experiment discussed in Section~\ref{sec:ablation_study}. Table~\ref{tab:ablated_transfer_dapt_advantage} presents the DAPT advantage values for each proficiency-based DAPT variant, while Table~\ref{tab:ablated_transfer_full_results} reports the complete raw transfer results from which these values are derived. DAPT advantage is computed using the same procedure described in Section~\ref{sec:cross-dataset_transfer_results}.

Overall, the ablated transfer results show that proficiency-targeted DAPT does not consistently improve cross-dataset transfer. In the IELTS to FCE direction, most DAPT advantage values remain negative across pretraining subsets and shot sizes, indicating that restricting DAPT to better-aligned EFCAMDAT subsets does not reliably reduce the transfer gap. The strongest exception is the B2--C1 subset at 200 shots, which shows small positive advantages for QWK, Spearman, and Pearson. In the opposite transfer direction, the results are somewhat more favorable, especially at 100 and 200 shots, where A1--A2, B1--B2, and B2--C1 DAPT produce positive RMSE advantages. However, the gains remain inconsistent across metrics, with B2--C1 showing the clearest positive pattern but not uniformly improving all metrics. These results suggest that proficiency alignment alone is insufficient to guarantee improved cross-test transferability.

\begin{table*}[t]
\centering
{\small
\setlength{\tabcolsep}{3pt}

\begin{minipage}{0.49\textwidth}
\centering
\textbf{IELTS $\rightarrow$ FCE}

\vspace{0.3em}
\begin{tabular}{llrrrr}
\toprule
Pretraining data & $n$ & RMSE & QWK & Sp. & Pe. \\
\midrule
Full EFCAMDAT & 50  & -0.369 & -0.015 & -0.013 & -0.012 \\
              & 100 & -0.395 & -0.058 & \textbf{0.031} & \textbf{0.025} \\
              & 200 & -0.270 & -0.103 & \textbf{0.005} & -0.012 \\
\midrule
A1--A2 only & 50  & -0.238 & -0.105 & -0.040 & -0.057 \\
            & 100 & -0.276 & -0.022 & \textbf{0.014} & \textbf{0.017} \\
            & 200 & -0.222 & -0.087 & -0.011 & -0.025 \\
\midrule
B1--B2 only & 50  & -0.338 & -0.072 & -0.039 & -0.052 \\
            & 100 & -0.387 & -0.137 & -0.034 & -0.046 \\
            & 200 & -0.330 & -0.082 & \textbf{0.023} & \textbf{0.010} \\
\midrule
B2--C1 only & 50  & -0.242 & -0.008 & -0.022 & -0.021 \\
            & 100 & -0.409 & -0.015 & \textbf{0.017} & \textbf{0.024} \\
            & 200 & -0.149 & \textbf{0.002} & \textbf{0.050} & \textbf{0.040} \\
\bottomrule
\end{tabular}
\end{minipage}
\hfill
\begin{minipage}{0.49\textwidth}
\centering
\textbf{FCE $\rightarrow$ IELTS}

\vspace{0.3em}
\begin{tabular}{llrrrr}
\toprule
Pretraining data & $n$ & RMSE & QWK & Sp. & Pe. \\
\midrule
Full EFCAMDAT & 50  & -0.066 & -0.324 & -0.032 & -0.043 \\
              & 100 & \textbf{0.010} & -0.082 & -0.020 & -0.023 \\
              & 200 & -0.025 & -0.125 & -0.025 & -0.040 \\
\midrule
A1--A2 only & 50  & -0.052 & -0.242 & -0.064 & -0.065 \\
            & 100 & \textbf{0.106} & \textbf{0.015} & -0.026 & -0.014 \\
            & 200 & \textbf{0.062} & -0.062 & -0.068 & -0.045 \\
\midrule
B1--B2 only & 50  & -0.040 & -0.313 & -0.061 & -0.062 \\
            & 100 & \textbf{0.068} & -0.004 & -0.008 & -0.020 \\
            & 200 & \textbf{0.073} & -0.055 & -0.035 & -0.032 \\
\midrule
B2--C1 only & 50  & -0.088 & -0.232 & 0.000 & -0.025 \\
            & 100 & \textbf{0.051} & \textbf{0.067} & \textbf{0.045} & \textbf{0.034} \\
            & 200 & \textbf{0.018} & \textbf{0.039} & \textbf{0.011} & -0.003 \\
\bottomrule
\end{tabular}
\end{minipage}
}

\caption{DAPT advantage in the proficiency-based cross-dataset transfer ablation. Each value compares the transfer-induced performance change of a BERT model adapted on a given EFCAMDAT subset against the corresponding non-adapted BERT baseline. Positive values indicate a more favorable transfer outcome for the adapted model. RMSE is treated inversely. Sp.=Spearman; Pe.=Pearson.}
\label{tab:ablated_transfer_dapt_advantage}
\end{table*}

\begin{table*}[t]
\centering
\scriptsize
\setlength{\tabcolsep}{4pt}
\renewcommand{\arraystretch}{1.08}
\begin{threeparttable}
\caption{Full results for the proficiency-based cross-dataset transfer ablation. Values in parentheses show the difference from the corresponding model's in-domain result.}
\label{tab:ablated_transfer_full_results}
\begin{tabular}{llrrrr}
\toprule
Pretraining data & $n$ & RMSE $\downarrow$ & QWK $\uparrow$ & Spearman $\uparrow$ & Pearson $\uparrow$ \\
\midrule

\multicolumn{6}{l}{\textbf{IELTS $\rightarrow$ FCE}} \\
\midrule
Full EFCAMDAT & In-domain & 2.101 & 0.598 & 0.652 & 0.678 \\
              & 50  & 2.557 (+0.456) & 0.429 (-0.169) & 0.490 (-0.162) & 0.515 (-0.163) \\
              & 100 & 2.433 (+0.332) & 0.483 (-0.115) & 0.596 (-0.056) & 0.616 (-0.062) \\
              & 200 & 2.271 (+0.170) & 0.510 (-0.088) & 0.590 (-0.062) & 0.605 (-0.073) \\
\addlinespace

A1--A2 only & In-domain & 2.177 & 0.581 & 0.647 & 0.665 \\
            & 50  & 2.502 (+0.325) & 0.322 (-0.259) & 0.458 (-0.189) & 0.457 (-0.208) \\
            & 100 & 2.390 (+0.213) & 0.502 (-0.079) & 0.574 (-0.073) & 0.595 (-0.070) \\
            & 200 & 2.299 (+0.122) & 0.509 (-0.072) & 0.569 (-0.078) & 0.579 (-0.086) \\
\addlinespace

B1--B2 only & In-domain & 2.064 & 0.609 & 0.662 & 0.679 \\
            & 50  & 2.489 (+0.425) & 0.383 (-0.226) & 0.474 (-0.188) & 0.476 (-0.203) \\
            & 100 & 2.388 (+0.324) & 0.415 (-0.194) & 0.541 (-0.121) & 0.546 (-0.133) \\
            & 200 & 2.294 (+0.230) & 0.542 (-0.067) & 0.618 (-0.044) & 0.628 (-0.051) \\
\addlinespace

B2--C1 only & In-domain & 2.143 & 0.550 & 0.636 & 0.653 \\
            & 50  & 2.472 (+0.329) & 0.388 (-0.162) & 0.465 (-0.171) & 0.481 (-0.172) \\
            & 100 & 2.489 (+0.346) & 0.478 (-0.072) & 0.566 (-0.070) & 0.590 (-0.063) \\
            & 200 & 2.192 (+0.049) & 0.567 (+0.017) & 0.619 (-0.017) & 0.632 (-0.021) \\
\addlinespace

No EFCAMDAT & In-domain & 2.316 & 0.571 & 0.677 & 0.684 \\
            & 50  & 2.403 (+0.087) & 0.417 (-0.154) & 0.528 (-0.149) & 0.533 (-0.151) \\
            & 100 & 2.253 (-0.063) & 0.514 (-0.057) & 0.590 (-0.087) & 0.597 (-0.087) \\
            & 200 & 2.216 (-0.100) & 0.586 (+0.015) & 0.610 (-0.067) & 0.623 (-0.061) \\

\midrule
\multicolumn{6}{l}{\textbf{FCE $\rightarrow$ IELTS}} \\
\midrule

Full EFCAMDAT & In-domain & 0.737 & 0.689 & 0.764 & 0.778 \\
              & 50  & 0.832 (+0.095) & 0.375 (-0.314) & 0.699 (-0.065) & 0.699 (-0.079) \\
              & 100 & 0.743 (+0.006) & 0.628 (-0.061) & 0.727 (-0.037) & 0.742 (-0.036) \\
              & 200 & 0.737 (+0.000) & 0.660 (-0.029) & 0.748 (-0.016) & 0.755 (-0.023) \\
\addlinespace

A1--A2 only & In-domain & 0.735 & 0.687 & 0.795 & 0.793 \\
            & 50  & 0.816 (+0.081) & 0.455 (-0.232) & 0.698 (-0.097) & 0.692 (-0.101) \\
            & 100 & 0.645 (-0.090) & 0.723 (+0.036) & 0.752 (-0.043) & 0.766 (-0.027) \\
            & 200 & 0.648 (-0.087) & 0.721 (+0.034) & 0.736 (-0.059) & 0.765 (-0.028) \\
\addlinespace

B1--B2 only & In-domain & 0.758 & 0.663 & 0.772 & 0.777 \\
            & 50  & 0.827 (+0.069) & 0.360 (-0.303) & 0.678 (-0.094) & 0.679 (-0.098) \\
            & 100 & 0.706 (-0.052) & 0.680 (+0.017) & 0.747 (-0.025) & 0.744 (-0.033) \\
            & 200 & 0.660 (-0.098) & 0.704 (+0.041) & 0.746 (-0.026) & 0.762 (-0.015) \\
\addlinespace

B2--C1 only & In-domain & 0.695 & 0.627 & 0.725 & 0.751 \\
            & 50  & 0.812 (+0.117) & 0.405 (-0.222) & 0.692 (-0.033) & 0.690 (-0.061) \\
            & 100 & 0.660 (-0.035) & 0.715 (+0.088) & 0.753 (+0.028) & 0.772 (+0.021) \\
            & 200 & 0.652 (-0.043) & 0.762 (+0.135) & 0.745 (+0.020) & 0.765 (+0.014) \\
\addlinespace

No EFCAMDAT & In-domain & 0.632 & 0.712 & 0.784 & 0.785 \\
            & 50  & 0.661 (+0.029) & 0.722 (+0.010) & 0.751 (-0.033) & 0.749 (-0.036) \\
            & 100 & 0.648 (+0.016) & 0.733 (+0.021) & 0.767 (-0.017) & 0.772 (-0.013) \\
            & 200 & 0.607 (-0.025) & 0.808 (+0.096) & 0.793 (+0.009) & 0.802 (+0.017) \\

\bottomrule
\end{tabular}
\begin{tablenotes}
\small
\item For RMSE, lower is better; therefore, a positive difference from in-domain means performance worsened, while a negative difference means performance improved. For QWK, Spearman, and Pearson, higher is better.
\end{tablenotes}
\end{threeparttable}
\end{table*}

\end{document}